\documentclass[letterpaper, 10 pt, conference]{llncs}  

\usepackage{graphics} 
\usepackage{epsfig} 
\usepackage{subfigure}
\usepackage{amsmath} 
\usepackage{epstopdf}
\usepackage{hyperref}
\usepackage{lipsum,amsmath,multicol}
\usepackage{float}
\usepackage{algorithm}
\usepackage{algpseudocode}

\title{\LARGE \bf
Chance constraint based multi agent navigation under uncertainty
}

\author{Bharath Gopalakrishnan$^{1}$, Arun Kumar Singh$^{2}$, Meha Kaushik$^{1}$, K. Madhava Krishna$^{1}$ and Dinesh Manocha$^{3}$  
\thanks{The research was supported }
\thanks{$^{1}$ RRC, IIIT-Hyderabad
      }%
\thanks{$^{2}$ NTU, Singapore
        }
\thanks{$^{3}$ University of North Carolina, Chappel Hill
        }%
}

\begin{document}

\maketitle


\begin{abstract}

We present Probabilistic Reciprocal Velocity Obstacle or PRVO  as a general algorithm for navigating multiple robots under perception and motion uncertainty. PRVO is defined as the space of velocities that ensures dynamic collision avoidance between a pair of robots with a specified probability. Our approach is based on defining chance constraints over the inequalities defined by the deterministic Reciprocal Velocity Obstacle (RVO). The computational complexity of the proposed probabilistic RVO is comparable to the deterministic counterpart. This is achieved by a series of reformulations where we first substitute the computationally intractable chance constraints with a family of surrogate constraints and then adopt a time scaling based solution methodology to efficiently characterize their solution space. Further, we also show that the solution space of each member of the family of surrogate constraints can be mapped in closed form to the probability with which the original chance constraints are satisfied and thus consequently to probability of collision avoidance. We validate our formulations through numerical simulations where we highlight the importance of incorporating the effect of motion uncertainty and the advantages of PRVO over existing formulations which handles the effect of uncertainty by using conservative bounding volumes.

\end{abstract}

\section{Introduction}
Ensuring collision free navigation of multiple robots between given start and goal positions is an important problem in robotics, swarm and crowd simulation. One commonly used approach to solve this problem performs local collision avoidance between multiple robots using velocity obstacles (VO) \cite{vo}, \cite{rvo}, \cite{mora}. This involves  repeatedly solving a set of non-convex inequalities to characterize the space of collision free velocities available to each robot at a given instant. Some faster techniques use a conservative formulation and reduce the local velocity computation to a linear programming problem \cite{orca}. 

Most algorithms for  multi-robot navigation focus primarily on the deterministic setting where it is assumed that the robot can perfectly estimate the states of the neighbouring robots and execute the computed avoidance maneuver without any errors. However, in reality both  motion and state estimation associated with each robot are imprecise and it is important to take the uncertainty in consideration. Thus, for robust implementation of multi robots in a particular task, it is imperative to make the transition to the probabilistic domain, where we explicitly consider the uncertainty in the system while computing the avoidance velocities.
As one would expect, the complexity of the probabilistic variant tends to be much  higher than the deterministic counterpart, as now the reasoning shifts from just \emph{collision avoidance} to\emph{ probability of collision avoidance}. To be more precise, in the probabilistic setting, it is required to map each velocity in a given set to the resulting probability of collision avoidance. Alternatively, one needs to obtain the characterization for the space of velocities for a given probability of collision avoidance.



In this paper, we present a novel algorithm for multi-robot collision avoidance which explicitly takes into account, the perception and motion uncertainty of each robot. Our approach combines ideas from velocity obstacle based multi-robot collision avoidance and robust control specifically robust Model Predictive Control (MPC) . In particular, we borrow the concept of chance constraints which are used in robust control to ensure constraint satisfaction under uncertainty \cite{mpc1}. These chance constraints ensure that the probability of a constraint being satisfied is greater than a specified threshold. In our approach, we formulate chance constraints over the conditions defined by the deterministic Reciprocal Velocity Obstacle (RVO) \cite{rvo} and characterize  the the resulting set of inequalities as Probabilistic Reciprocal Velocity Obstacle or PRVO.



We also provide a computationally efficient approach for solving these complex set of inequalities. To this end, we build upon our recent series of works  \cite{iros15}, \cite{cdc15}, where  chance constraints are substituted with a more tractable family of surrogate constraints. The solution space of each member of the family can be  mapped in closed form to the probability with each the original chance constraints are satisfied. Finally, a time scaling based methodology can be adopted to efficiently solve the surrogate constraints. In this paper, we extend the approach of \cite{iros15}, \cite{cdc15} to multiple decision making robots by expanding our model to incorporate the uncertainty that the robot would exhibit while executing the computed avoidance maneuver. 

\subsection{Summary of Main Results}
On the computational side, we show that the complexity of the PRVO is comparable to its deterministic counterpart. This is primarily achieved by the use of surrogate constraints and a time scaling based methodology for their solution. Further, we show that the previous developed Cantellis inequality based bounds which relates solution space of surrogate constraints to the probability of collision avoidance performs well in case of maneuvering targets as well. On the practical side, we highlight the importance of explicitly including the uncertainty associated with execution of avoidance maneuvers. Moreover, the advantage of PRVO over existing works which incorporates the effect of uncertainty by enlarging the size of the robot \cite{hrvo},\cite{hrvo_prvo}, \cite{calu}, \cite{cocalu} is also presented.

\subsection{Layout of the Paper}

The rest of the paper is organized as follows. Section \ref{rel} contrasts the proposed formulation with the existing works. Section \ref{notation} summarizes the notations used in the paper. Section \ref{pre} presents the collision avoidance conditions as modeled Reciprocal Velocity Obstacle (RVO) in the deterministic setting. Section \ref{stochastic} introduces the model of the uncertainty used in the current work followed by the introduction of chance constraints over the inequalities defined by RVO. Section \ref{timescale} presents a time scaling based methodology for efficient characterization of the solution space of the surrogate substitution of the chance constraints. Section \ref{analysis} presents an analysis of computational complexity, probability bounds and motivations for choosing RVO as the basis for construction of chance constraints as compared to its convex approximation ORCA. The simulation results are presented in section \ref{res}

\section{Related Work}\label{rel}
In this section, we contrast our proposed formulation with the existing works in terms of uncertainty model, methodology for formulating and solving dynamic collision avoidance under uncertainty in the context of both single robot and multi-robot navigation.

\subsection{Uncertainty Model}
Our methodology of modeling uncertainty in the system is similar to that presented in works like \cite{Kluge}, \cite{Kim}, \cite{laugier1}, \cite{laugier2}, \cite{likhachev}, \cite{trautman_unfreezing}, \cite{royauro}, \cite{luders}, \cite{hrvo}, \cite{calu}, \cite{cocalu}. in the sense that  uncertainty at any time instant is represented as a normal random variable with a particular mean vector and covariance matrix.  However, among these, \cite{Kluge}-\cite{royauro} consider the case of a single robot moving among non-responsive dynamic obstacles and incorporates the effect of only perception uncertainty while treating the robot's motion through deterministic models. Algorithms presented in \cite{luders} considers the effect of both perception and motion uncertainty but only in the context of dynamic collision avoidance of a single robot. In \cite{hrvo}, each robots motion uncertainty is taken into consideration while constructing the RVO. However, it does not consider the uncertainty in the error between the computed and executed avoidance maneuver,or in short the actuation uncertainty.

\subsection{Dynamic Collision Avoidance Constraints under Uncertainty}

At a conceptual level, our modeling of dynamic collision avoidance under uncertainty is similar to works like \cite{Kluge}-\cite{luders} in the sense that we are also primarily concerned with relating each velocity in  some  given set to a probability of collision avoidance. The differentiating factor however, with these cited works lies in the technical approach followed and  in the fact that we deal with multiple maneuvering robots. \cite{Kluge}-\cite{luders} relies on sampling different velocities, using them  to construct robot's trajectory over a short horizon and inferring probability of collision with respect to them. However, such sampling based approaches are difficult to extend for multiple robots due increase in complexity of the search space. In contrast, we propose a systematic decomposition of probabilistic collision avoidance constraints into simpler forms eventually leading to a close form approximation of the solution space. 

The probabilistic approach followed in the current proposed work is also very different from those presented in \cite{hrvo}, \cite{calu}, \cite{cocalu} which account for uncertainty in dynamic collision avoidance by expanding the size of the velocity and radius of the robot, depending on the level of uncertainty in the system. In principle, this is equivalent to drawing a lot of samples from position, size and velocity uncertainty ellipse and writing collision avoidance with respect to each of them. Although very simple, this approach suffers from the drawback that it is blind to the probability of samples drawn and thus, samples which are closer to the mean are given same importance as samples farther from the mean. At the implementation level, one would have the same collision avoidance constraints with respect to both the samples. As we show later, this leads to a very conservative solution space.

%

%
%

\section{Symbols and Notations}\label{notation}
We used bold faced small case letters with superscripts to describe vectors associated with a particular robot. For example, the position and velocity of robot $i$ is represented as $\textbf{p}^i=(p_x, p_y)$ and $\textbf{v}^i=(v_x, v_y)$. We use $f^{RVO^i_j}$ to represent the collision avoidance conditions computed through the concept of RVO. The symbol $\textbf{v}^i_{rvo}$ represents collision avoiding velocities computed using the deterministic RVO algorithm. We use $\mu$ and $\sigma^2$ with suitable subscripts and superscripts to represent mean and variance of vectors or functions respectively. In section \ref{timescale}, we use additional superscript $"s"$ to denote time scaled variants of vectors, functions or constraints. 

\section{Pre-Requsitie: Deterministic  Velocity and Reciprocal Velocity Obstacle}\label{pre}
In this section, we briefly review the concept of Reciprocal Velocity Obstacle (RVO). We do not go into  details, rather present the general algebraic form for the collision avoidance constraints defined by RVO. For details, the readers are requested to refer to \cite{rvo}. We consider disk shaped robots each modeled as following single integrator system in 2D Euclidean $X-Y$ space

\begin{equation}
\dot{x}^i = v_x^i \Rightarrow x^i(t+\Delta t) = x^i(t)+v_x^i \Delta t, \dot{y}^i = v_y^i \Rightarrow y^i(t+\Delta t) = y^i(t)+v_y^i \Delta t 
\label{evol_stline}
\end{equation}

In (\ref{evol_stline}), $v_x^i$ and $v_y^i$ respectively represents the $x$ and $y$ component of the velocity of the $i^{th}$ robot. It is clear from (\ref{evol_stline}) that the trajectory of each robot is assumed to be composed of piece-wise straight line segments. 


Now, consider a collision scenario where two robots with radius $R^i$ and $R^j$ are moving with constant velocities $\textbf{v}^i$ and $\textbf{v}^j$. The space of such velocities $\textbf{v}^i_{rvo}$ which allow robot $i$ to come out of the collision course with robot $j$ can be characterized by the following set of inequalities derived from the concept of RVO \cite{rvo}.

\begin{equation}
f^{RVO^i_j} (\textbf{p}^j, \textbf{v}^j)\geq 0 
\label{RVO_comp}
\end{equation}

\begin{equation}\nonumber
f^{RVO^i_j} (\textbf{p}^j, \textbf{v}^j) = \Vert \textbf{r}^{ij} \Vert^2 - \frac{(\textbf{r}^{ij})^T(2\textbf{v}^i_{rvo}-\textbf{v}^i-\textbf{v}^j)}{\Vert 2\textbf{v}^i_{rvo}-\textbf{v}^i-\textbf{v}^j\Vert^2}-R^2
\end{equation}

It is straightforward to observed that (\ref{RVO_comp}) is a non-convex quadratic with respect to $\textbf{v}^i_{rvo}$ and thus, computing a characterization of collision free velocities automatically becomes a challenging problem. 

At this point, it is worth pointing out that the in the deterministic setting it is assumed that each robot can estimate its current state and the state of the other robot perfectly and thus, can construct inequalities (\ref{RVO_comp}) exactly.  In the subsequent sections, we relax this assumption thereby extending RVO to the probabilistic domain.


\section{RVO in the Probabilistic Domain}\label{stochastic}
Let us start by representing the current trajectory of each robot as the following random variables.

\begin{eqnarray}
\textbf{p}^i \approx N(\mu^i_{\textbf{p}}, (\sigma^i_{\textbf{p}})^2), \hspace{0.2cm} \textbf{p}^j \approx N(\mu^j_{\textbf{p}}, (\sigma^j_{\textbf{p}})^2) \label{pos_uncert}\\
\textbf{v}^i \approx N(\mu^i_{\textbf{v}}, (\sigma^i_{\textbf{v}})^2), \hspace{0.2cm} \textbf{v}^j \approx N(\mu^j_{\textbf{v}}, (\sigma^j_{\textbf{v}})^2),\label{vel_uncert}
\end{eqnarray}

where $\mu^i_{\textbf{p}}, (\sigma^i_{\textbf{p}})^2, \mu^i_{\textbf{v}}, (\sigma^i_{\textbf{v}})^2$ and similarly others represents position and velocity level mean and variances, respectively. In the context of two robot collision scenario considered in the previous section, equations (\ref{pos_uncert}) and (\ref{vel_uncert}) model the fact that robot $i$  has some uncertainty in the estimate of its current state and the state of the robot $j$. Although, we have assumed a Gaussian form, our approach  can be easily extended to incorporate other representations as well and we discuss that briefly, later on in the paper.

Similarly, let us assume that each robot has an imperfect actuation and thus there is an inherent noise between the commanded and actual velocity. This noise would result in some error between the avoidance maneuver that the robot intends to perform and actually performs. Moreover, this error itself would be a random variable. In the context of RVO, we account for this uncertainty associated with avoidance maneuver by assuming that $\textbf{v}^i_{rvo}$ is drawn from a distribution. In other words, it is modeled as the following Gaussian random variable

\begin{equation}
\textbf{v}^i_{rvo} \approx N({\textbf{v}^i_{rvo}}, (\sigma^i_{\textbf{v}_{rvo}})^2),
\label{motion_uncert}
\end{equation}

The above equation models the fact that when the robot commands a velocity $\textbf{v}^i_{rvo}$, the executed velocity can correspond to any sample drawn from a Gaussian distribution whose mean is the commanded velocity and the variance is some constant $(\sigma^i_{\textbf{v}_{rvo}})^2$

Now, in light of definitions (\ref{pos_uncert})-(\ref{motion_uncert}), $f^{RVO^i_j}$ becomes a multivariate function of random variables and thus, consequently a random variable itself. Thus, mathematically,  (\ref{RVO_comp}) does not make sense. Instead, a more well defined alternative would be to consider the following inequality

\begin{equation}
P(f^{RVO^i_j} (\textbf{p}^j, \textbf{v}^j)\geq 0)\geq \eta .
\label{PRVO}
\end{equation}

Constraint (\ref{PRVO}) ensures that the probability of RVO based collision avoidance condition (\ref{RVO_comp}) being satisfied is greater than some lower bound $\eta$. It is straightforward to note that (\ref{PRVO}) infact defines the space of velocities $\textbf{v}^i_{rvo}$ for robot $i$ which ensures satisfaction of RVO constraints with atleast probability $\eta$ for the given robot $j$ trajectory parameters $\textbf{p}^j$ and $\textbf{v}^j$. We define (\ref{PRVO}) as probabilistic reciprocal velocity obstacle or PRVO.

Constraints having the general form as that of (\ref{PRVO}) are popularly known as "chance constraints" in the robust control literature \cite{mpc1},  and in general are computationally intractable \cite{chance1}. The primary difficulty lies in computing the analytical form for the chance constraints. One notable exception exists in the case when the random variables in consideration have Gaussian distribution and the chance constraints are defined over affine inequalities \cite{boyd}. In such cases, efficient convex approximations for the chance constraints can be derived. However, as stated earlier, $f^{RVO^i_j}$ is non-convex quadratic in terms of random variables and thus the techniques proposed in \cite{boyd} is not applicable in our case. In the next section, we present a novel substitution for (\ref{PRVO}), which exploits the fact that although it is intractable to obtain the analytical form for left hand side of (\ref{PRVO}), it is relatively straightforward to obtain symbolic expressions for expectation and variance for $f^{RVO^i_j}$.

\subsection{Expectation and Variance of $f^{RVO^i_j}(.)$}

Expectation of a multivariate function $g$ in terms of variables $z_1,z_2...z_n$  is given by the following expression. 

 \small
 \begin{eqnarray}\label{expec}
 E[g(z_1,z_2...z_n)]= \\\nonumber \int_{-\infty}^{\infty}...\int_{-\infty}^{\infty}g(z_1,z_2...z_n)h(z_1,z_2...z_n)dz_1dz_2..dz_n
 \end{eqnarray}
 \normalsize

Using (\ref{expec}), the expectation of $f^{RVO^i_j}(.)$ can be obtained in the following manner.

\small
\begin{equation}
E[f^{RVO^i_j}] = \mu_{f^{RVO^i_j}}= \int_{-\infty}^{\infty}...\int_{-\infty}^{\infty}f^{RVO^i_j}(.)P(.) dx^idy^idx^jdy^jd{v_x}^id{v_y}^id{v_x}^jd{v_y}^j d({v}^i_{rvo})_x, d({v}^i_{rvo})_y
\label{expec2}  
\end{equation}
\normalsize

In (\ref{expec2}) $P(.)$ represents the joint probability distribution of the random variables $(x^i, y^i, x^j, y^j, {v_x}^i, {v_y}^i, {v_x}^j, {v_y}^j, ({v}^i_{rvo})_x, ({v}^i_{rvo})_y)$. Integral (\ref{expec2}) can be computed symbolically using packages like MATHEMATICA \cite{mathematica} and can be eventually reduced to a quadratic polynomial in terms of $\textbf{v}^i_{rvo}$.

We can proceed to use (\ref{expec2}) to compute the variance of $f^{RVO^i_j}(.)$ in the following manner. The right hand side of equality (\ref{var}), when computed symbolically, reduces to a fourth order polynomial in terms of $\textbf{v}^i_{rvo}$

\small
\begin{equation}
(\sigma_{f^{RVO^i_j}})^2= {E[(f^{RVO^i_j}-E[f^{RVO^i_j}])^2]} 
\label{var}
\end{equation}
\normalsize

Equations having the general form similar to (\ref{expec2}) and (\ref{var}) were introduced in our earlier work \cite{iros15}. However, in contrast to our earlier formulations, (\ref{expec2}) and (\ref{var}) is more complex as they depend on additional random variables pertaining to motion uncertainty of the robot $i$.
%

\subsection{Approximations for PRVO}

Using (\ref{expec2}) and (\ref{var}), we can replace (\ref{PRVO}) with the following family surrogate constraints.

\begin{equation}
\mu_{f_{RVO^i_j}}-k\sigma_{f_{RVO^i_j}} \geq 0, \hspace{0.2cm}k\geq 0
\label{expecvariance}
\end{equation}

As shown in our earlier work \cite{iros15}, inequality (\ref{expecvariance}) represents a strip of width $\mu_{f_{RVO^i_j}}- k\sigma_{f_{RVO^i_j}}$ from the distribution of $f_{RVO^i_j}$. Thus, solving (\ref{expecvariance}) with increasing value of $k$ ensures increasingly larger part of the distribution of  $f_{RVO^i_j}$ to be above zero. This in turn  leads to increasingly safer velocities. To put it mathematically, satisfaction of (\ref{expecvariance}) ensures satisfaction of original chance constraints (\ref{PRVO}) with a lower bound probability dependent on the value of $k$, i.e $\eta \geq c(k)$ for some positive monontonic  function $c(k)$. In \cite{iros15}, we derived the following mapping based on Cantellis inequality, thus providing a closed form mapping between the solution space of each member of the family of constraints (\ref{expecvariance}) and the probability with which the original chance constraints are satisfied.

\begin{equation}
\eta\geq \frac{k^2}{1+k^2}
\label{Cantelli}
\end{equation}

\section{Time Scaling based solution of Surrogate Constraints}\label{timescale}

As stated earlier, (\ref{expec2}) and (\ref{var}) are respectively quadratic and quartic in terms of variable $\textbf{v}^i_{rvo}$. Thus, (\ref{expecvariance}) represents a non-convex polynomial inequality, which in general is computationally intractable. In this section, we present a time scaling based framework for approximating the solution space of (\ref{expecvariance}). The basic idea is simple. We first compute the space of velocities $\textbf{v}^i_{rvo}$ which the robot can reach by just changing the time scale of the current trajectory. Importantly, we show that this solution space can be characterized in closed form to obtain a set of formuale. Further, evaluating these formulae over multiple paths gives a good approximation of the complete solution space of (\ref{expecvariance}).

\subsection{Time Scaled Variant of Surrogate Constraints}
Changing the time scale of a trajectory from $t$ to $\tau$ does not alter the geometric path but results in following change in the velocity profile

\begin{equation}
\textbf{v}(\tau) = \textbf{v}(t)s, \hspace{0.2cm} s = \frac{dt}{d\tau}
\label{timedef}
\end{equation}

In (\ref{timedef}), $\frac{dt}{d\tau}$ is called the scaling function and decides the mapping  between the time scales. Now, with slight abuse of notation, let us assume that the robot $i$ is moving along a straight line trajectory characterized by a velocity $\textbf{v}^i$. Let us denote by following the space of collision avoiding velocities that the robot can achieve by just changing the time scale of the current trajectory

\begin{equation}
^s\textbf{v}^i_{rvo} = s\textbf{v}^i
\label{solpath}
\end{equation}

Substituting (\ref{solpath}) into (\ref{expec2}) and (\ref{var}) results in the following time scaled variant of the expectation and variance of $f^{RVO^i_j}$.

\begin{equation}
\mu_{^sf^{RVO^i_j}} = as^2+bs+c
\label{expec4}
\end{equation}

\begin{equation}
\sigma_{^sf^{RVO^i_j}}= \sqrt{E[(^sf^{RVO^i_j}-E[^sf^{RVO^i_j}])^2]} = \sqrt{ds^4+es^3+fs^2+gs+h}
\label{var4}
\end{equation}

Where, $a(.), b(.)...h(.)$ are function of parameters of distribution of random variables i. e, $\mu^i_{\textbf{p}}, (\sigma^i_{\textbf{p}})^2, \mu^j_{\textbf{p}}, (\sigma^j_{\textbf{p}})^2, \mu^i_{\textbf{v}}, (\sigma^i_{\textbf{v}})^2$ etc. Please note the additional superscript "$s$" representing that we are now dealing a time scaled variant of $\mu_{f_{RVO^i_j}}$ and $\sigma_{f_{RVO^i_j}}$ derived in (\ref{expec2}) and (\ref{var})

Now, we perform one final simplification of (\ref{var4}) to obtain quadratic approximation based on second order Taylor series expansion around a point $s^*$

\begin{equation}
\sigma_{^sf^{RVO^i_j}}= \sigma_{^s{^*}f^{RVO^i_j}}+ \sigma_{^s{^*}f^{RVO^i_j}}^{'}(s-{s^*})+\sigma_{^s{^*}f^{RVO^i_j}}^{''}\frac{(s{^*}-{s^*})^2}{2}
\label{quadapprox1}
\end{equation}

Using (\ref{expec4}) and (\ref{quadapprox1}), the final form of the time scaled variant of (\ref{expecvariance}) can be obtained in the following manner.

\begin{equation}
as^2+bs+c - k ( \sigma_{^s{^*}f^{RVO^i_j}}+ \sigma_{^s{^*}f^{RVO^i_j}}^{'}(s-{s^*})+\sigma_{^s{^*}f^{RVO^i_j}}^{''}\frac{(s-{s^*})^2}{2} )\geq 0
\label{timescalevariant}
\end{equation}

It can be observed that (\ref{timescalevariant}) is a single variable quadratic inequality and thus, its solution space can be characterized in  closed form \cite{iros14}. Extending (\ref{timescalevariant}) to say $n$ robots is also straightforward as in that case we would have $n$ single variable quadratic inequalities, the solution space of which can again be characterized in closed form. The inequality (\ref{timescalevariant}) can be constructed along multiple paths to obtain a good characterization of the complete solution space. 

It is worth pointing out that there are various heuristics for choosing the point $s^*$. One of them which has been shown to result in low approximation errors is to choose $s^*$ from the solution space of $\mu_{^sf^{RVO^i_j}} \geq 0$ \cite{iros15}. It is easy to note that solving it is similar to solving (\ref{timescalevariant}).

\section{Analysis of PRVO}\label{analysis}
\subsection{Computational Complexity of RVO and PRVO}.
The collision avoidance through deterministic RVO for a  pair of robots takes the form of non-convex quadratic inequality (\ref{RVO_comp}). In contrast, the surrogate constraints (\ref{expecvariance}) which approximates the PRVO constraints (\ref{PRVO}) is a quartic inequality. Thus, on the surface PRVO seems significantly more complicated than RVO. However, the time scaling based reformulations discussed in section \ref{timescale} does provide a significant simplification. Essentially as explained in section \ref{timescale}, solving the surrogate constraints and consequently PRVO has been reduced to generating multiple candidate trajectories and evaluating the solution space of (\ref{timescalevariant}). Generating multiple candidate trajectories is simple; during implementation, we randomly generate some velocity samples and then use them to construct straight line trajectories. Solving (\ref{timescalevariant}) is straightforward and as explained can be infact done in closed form. Although it is difficult to derive an bounds on the number of samples, we employ some heuristics and often recover a solution space with mostly one or two samples. Thus, it can be concluded that kind of surrogate constraints that we have proposed coupled with time scaling based solution makes computational complexity of PRVO atleast comparable to RVO.

%
%

%

\subsection{PRVO Vs PORCA}
 In the deterministic setting, the collision avoidance conditions modeled through ORCA for a pair of robots can be written as the following linear inequality for some $z_1$ and $z_2$ which are a function of robot $j$ trajectory parameters.

\begin{equation}
f^{ORCA^i_j} = z_1\textbf{v}^i_{orca}-z_2 \geq 0
\label{orca}
\end{equation}

In (\ref{orca}), $\textbf{v}^i_{orca}$ is the collision avoidance velocity as modeled through ORCA. Now, since ORCA is an approximation of RVO, the space of collision avoidance velocities characterized by (\ref{orca}) is much smaller than that characterized by RVO (\ref{RVO_comp}). Now let us consider chance constraints defined over (\ref{orca}) and its surrogate constraints given by \cite{boyd}

\begin{eqnarray}
P(f^{ORCA^i_j}\geq 0)\geq \eta, \hspace{0.2cm} z_1 \approx N(\mu_{z_1}, \sigma_{z_1})\label{orca_chance}\\
\Rightarrow z_1\textbf{v}^i_{orca}-z_2 -\eta^{\frac{1}{2}}((\textbf{v}^i_{orca})^2-2z_2\mu_{z_1}+z_2^2)^{\frac{1}{2}}\geq 0 \label{orca_surrogate}
\end{eqnarray}

Ineqaulity, \ref{orca_surrogate} represents a convex second order cone constraint. Moreover, the first two terms in \ref{orca_surrogate}, is conjunction is nothing but the deterministic ORCA (\ref{orca}), while the third term is non-negative. It is clear from (\ref{orca_surrogate}) that chance constraints essentially boils down to obtaining a even smaller subset from the already restrictive solution space of deterministic ORCA. Thus, at the implementation level, infeasibility of (\ref{orca_surrogate}) can be of concern.

In light of the above arguments, it obviously makes more sense to start with the larger solution space of RVO such that feasibility can still be ensured when the solution space shrinks because of the application of chance constraints.

\section{Results}\label{res}

The results presented in this section are grouped into following categories. (1) Validating that solving the surrogate constraints (\ref{expecvariance}), with increasing value of $k$ leads to satisfaction of original PRVO constraints (\ref{PRVO}) with increasing probability $\eta$. (2) Showing the importance of incorporating the motion uncertainty. (3) Comparing solution space computed through PRVO with that obtained from \cite{hrvo} and similar approaches.

\subsection{Validating Mapping between the Surrogate Constraints and the PRVO constraints }
\begin{figure*}[!t]
  \centering
    \subfigure[]{
\includegraphics[width= 5.5cm, height=4.5cm]{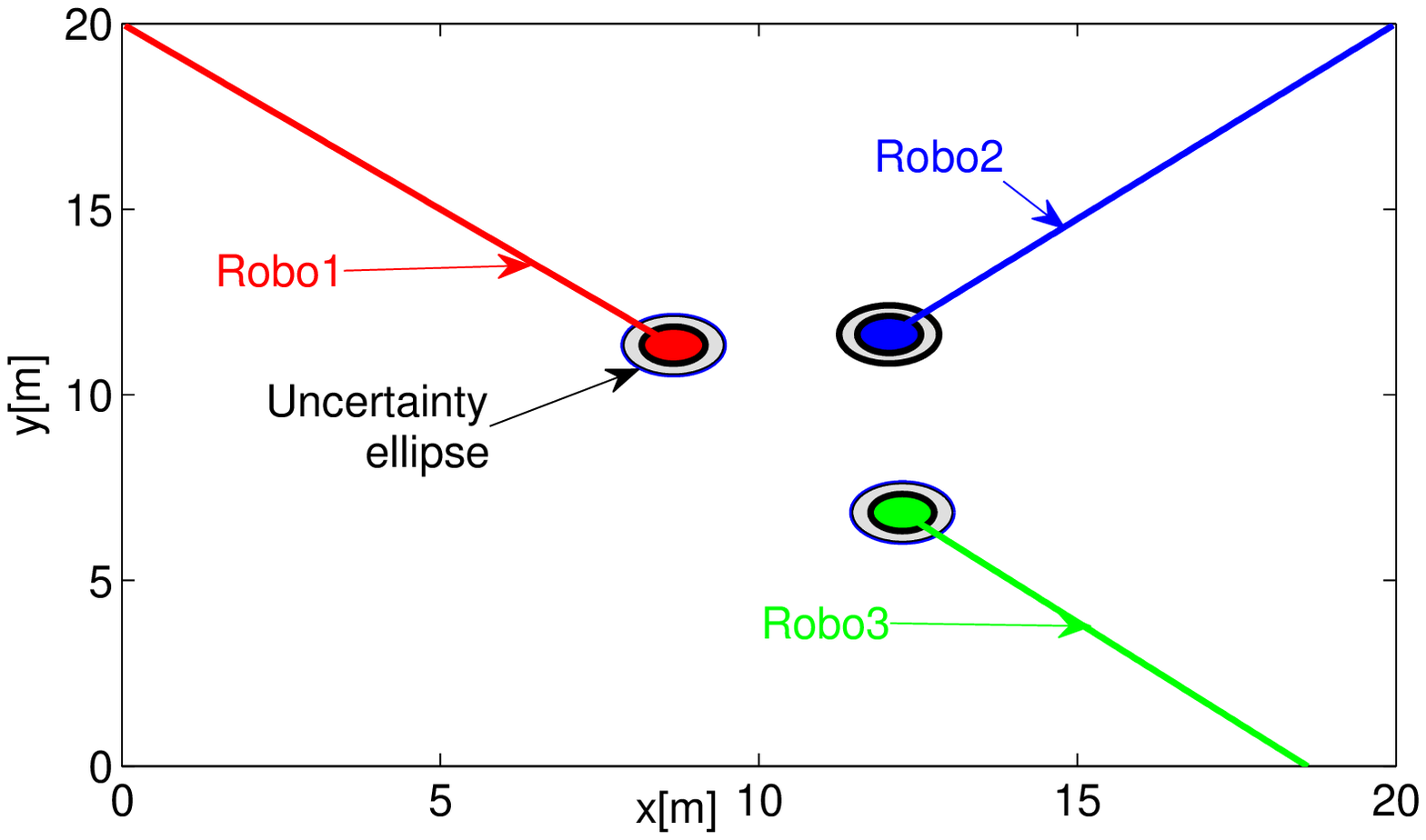}
        \label{3rob_coll}
        }\hspace{-0.4cm}
\subfigure[]{
\includegraphics[width= 5.5cm, height=4.5cm]{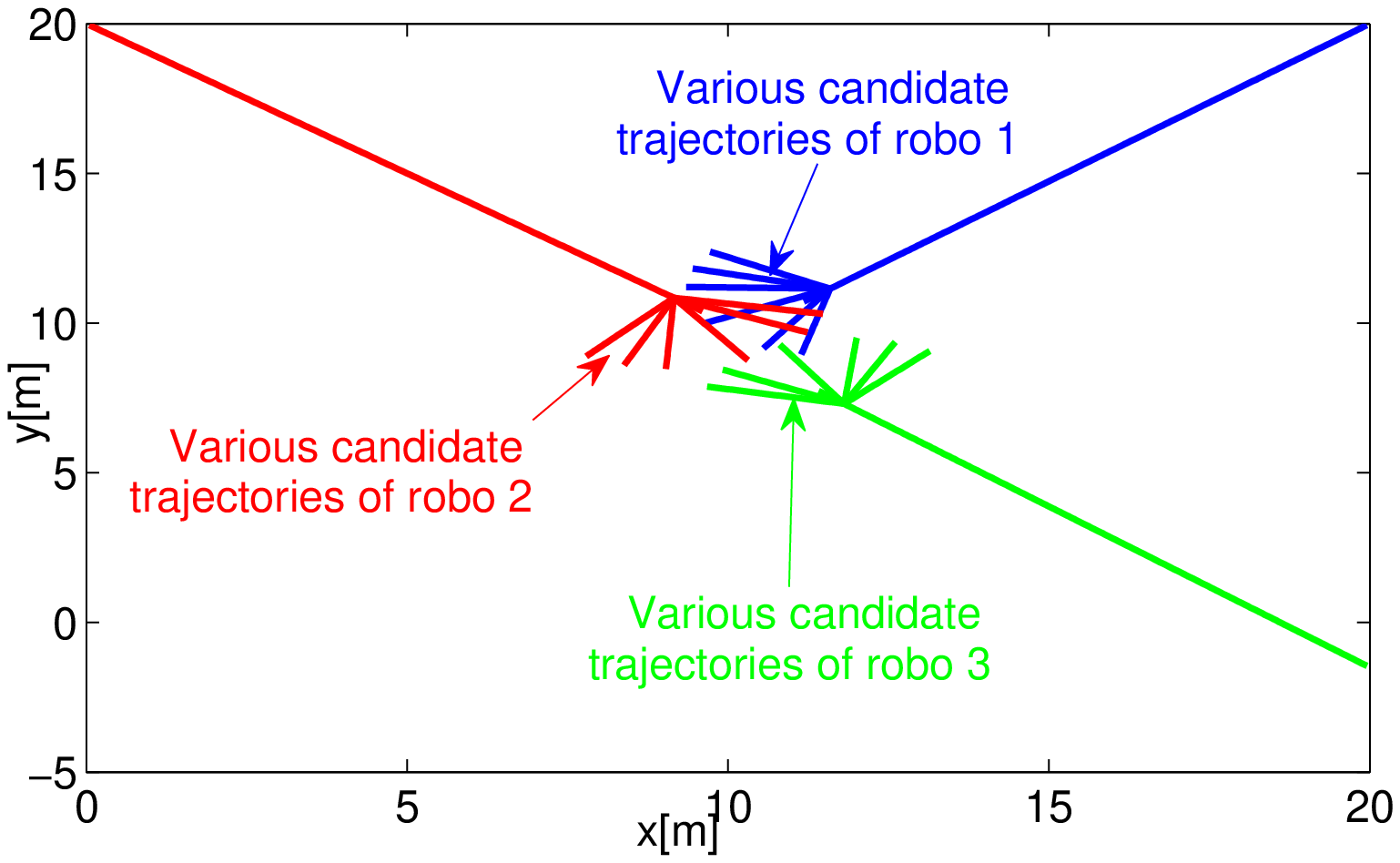}
        \label{candtraj1}
        }
\caption{(a): Collision scenario involving three robots. (b). For computing avoidance maneuvers, each robot generates various candidate trajectories and then solves the time scaled variant of the surrogate constraints (\ref{timescalevariant}) along them. Computing an avoidance maneuver would essentially mean choosing a particular candidate trajectory and a corresponding time scale for it.}
\end{figure*}

\begin{figure*}[!h]
  \centering
    \subfigure[]{
\includegraphics[width= 4.3cm, height=4.3cm]{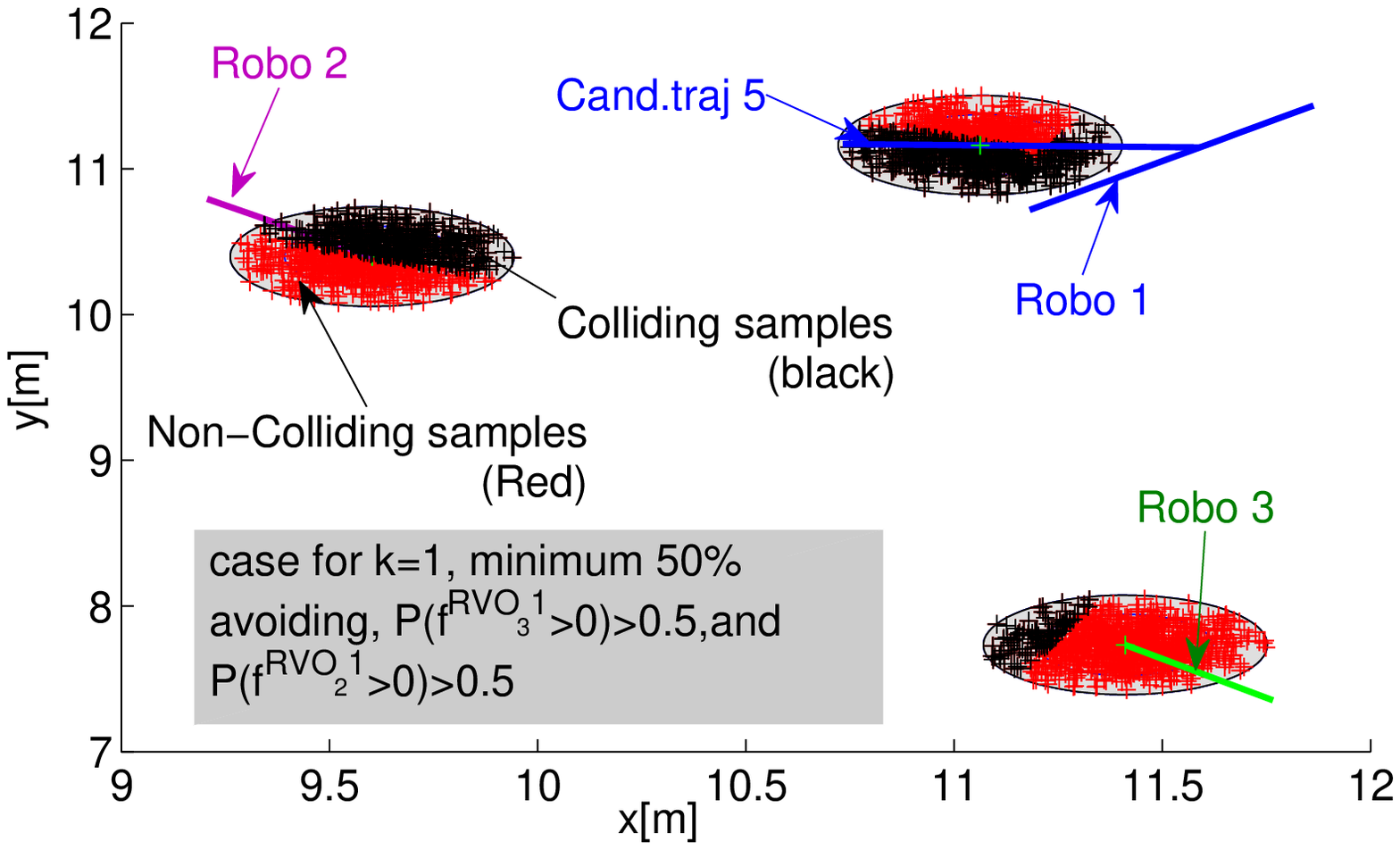}
        \label{k1rob1}
        }\hspace{-0.9cm}
\subfigure[]{
\includegraphics[width= 4.3cm, height=4.3cm]{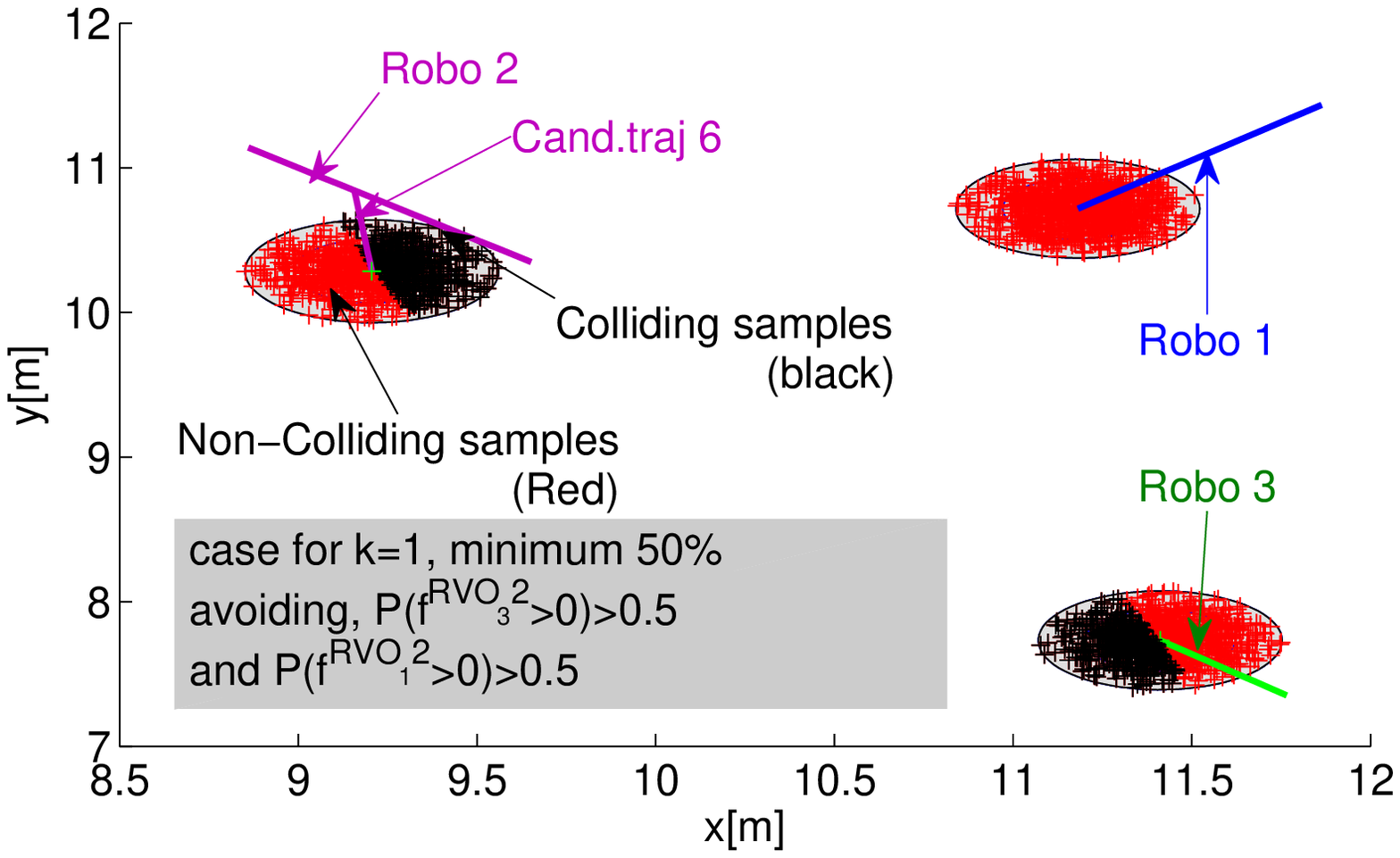}
        \label{k1rob2}
        }\hspace{-0.9cm}
        \subfigure[]{
\includegraphics[width= 4.3cm, height=4.3cm]{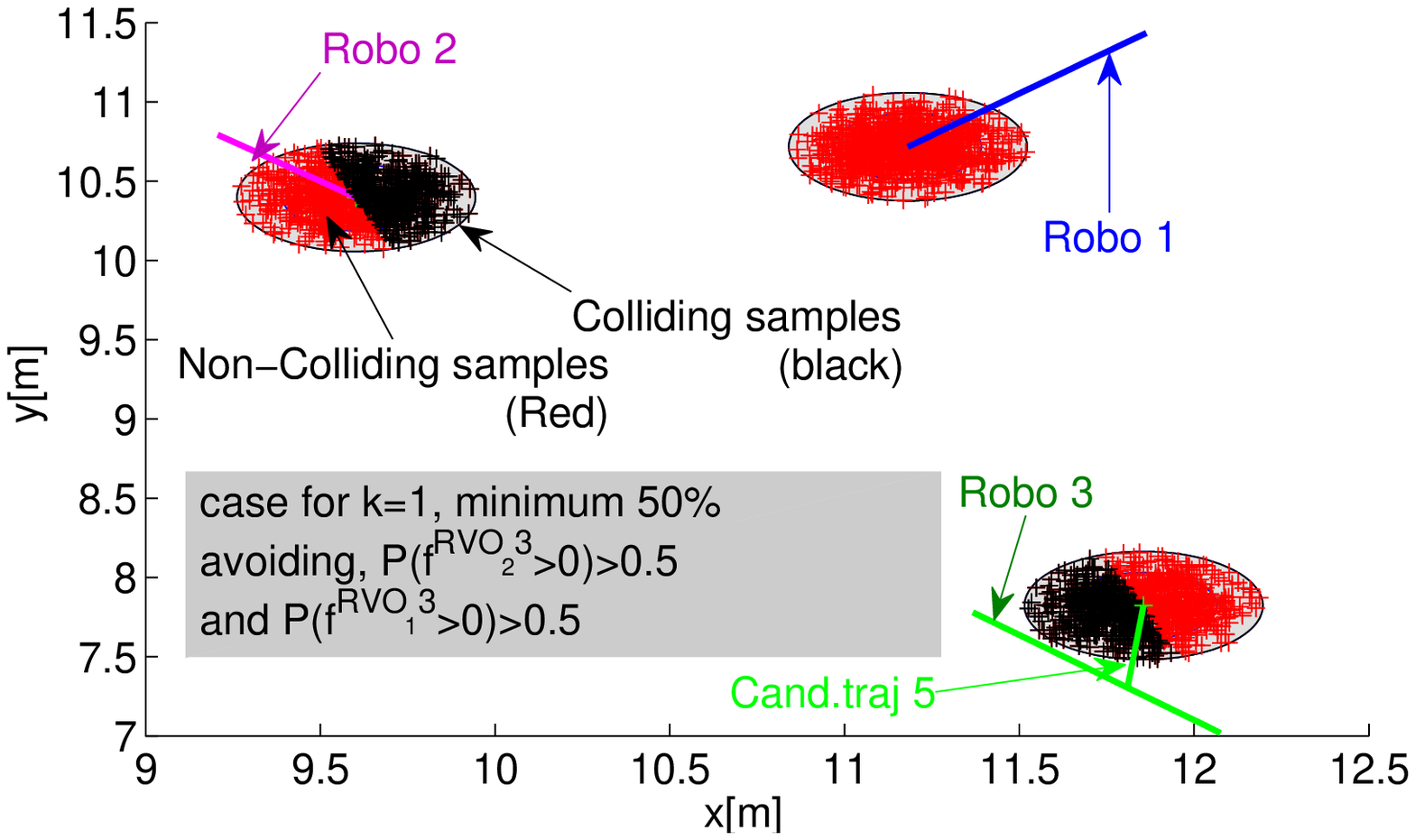}
        \label{k1rob3}
        }             
\caption{(a): Each robot chooses a candidate trajectory and then solves the time scaled variant of the surrogate constraints (\ref{timescalevariant}) along them for $k=1$. Based on the Cantellis bounds discussed in section 5.2, this should lead to satisfaction of PRVO constraints (\ref{PRVO}) with atleast probability 0.5. As can be seen from the figures, we validate that this is indeed the case. To elaborate further, we take various samples from the position and velocity uncertainty ellipse of each robots and evaluate what percentage of these samples lead to satisfaction of RVO constraints (\ref{RVO_comp}), and this is shown in the form of black and red samples in each uncertainty ellipse, where black samples indicate that the RVO constraints are not satisfied, and the red samples indicate that the RVO constraints are satisfied  }.
\end{figure*}

\begin{figure*}[!h]
  \centering
    \subfigure[]{
\includegraphics[width= 4.3cm, height=4.3cm]{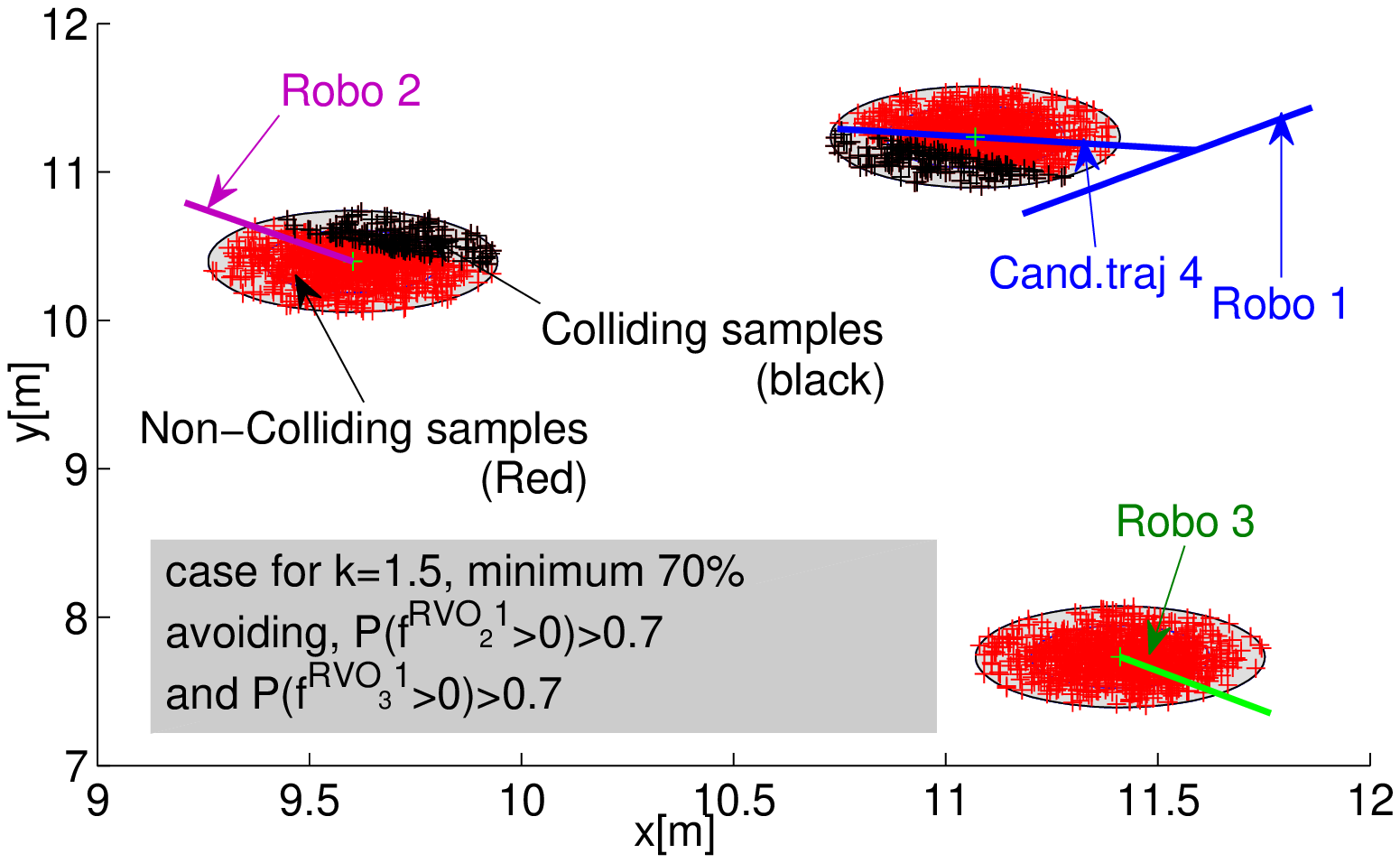}
        \label{k2rob1}
        }\hspace{-0.9cm}
\subfigure[]{
\includegraphics[width= 4.3cm, height=4.3cm]{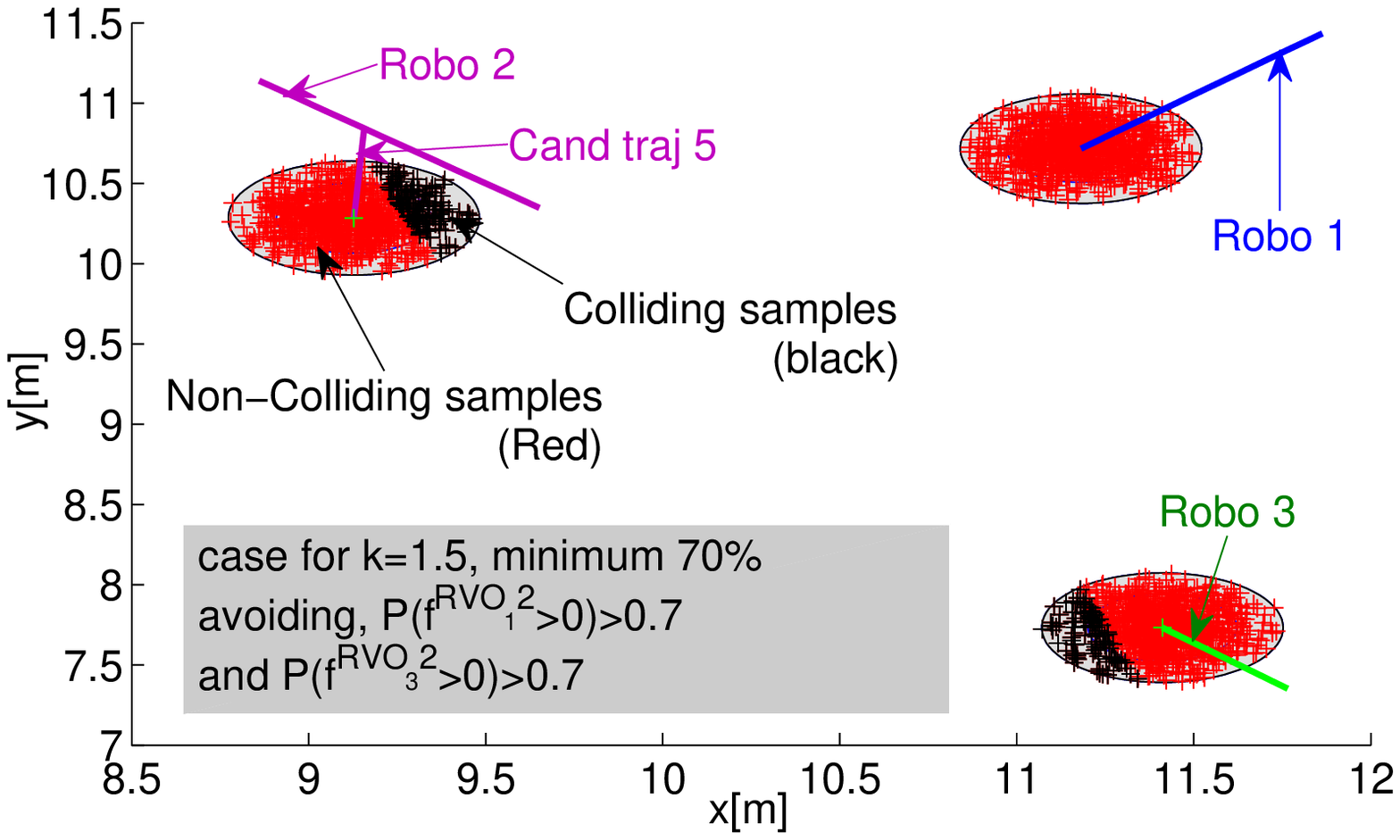}
        \label{k2rob2}
        }\hspace{-0.9cm}
        \subfigure[]{
\includegraphics[width= 4.3cm, height=4.3cm]{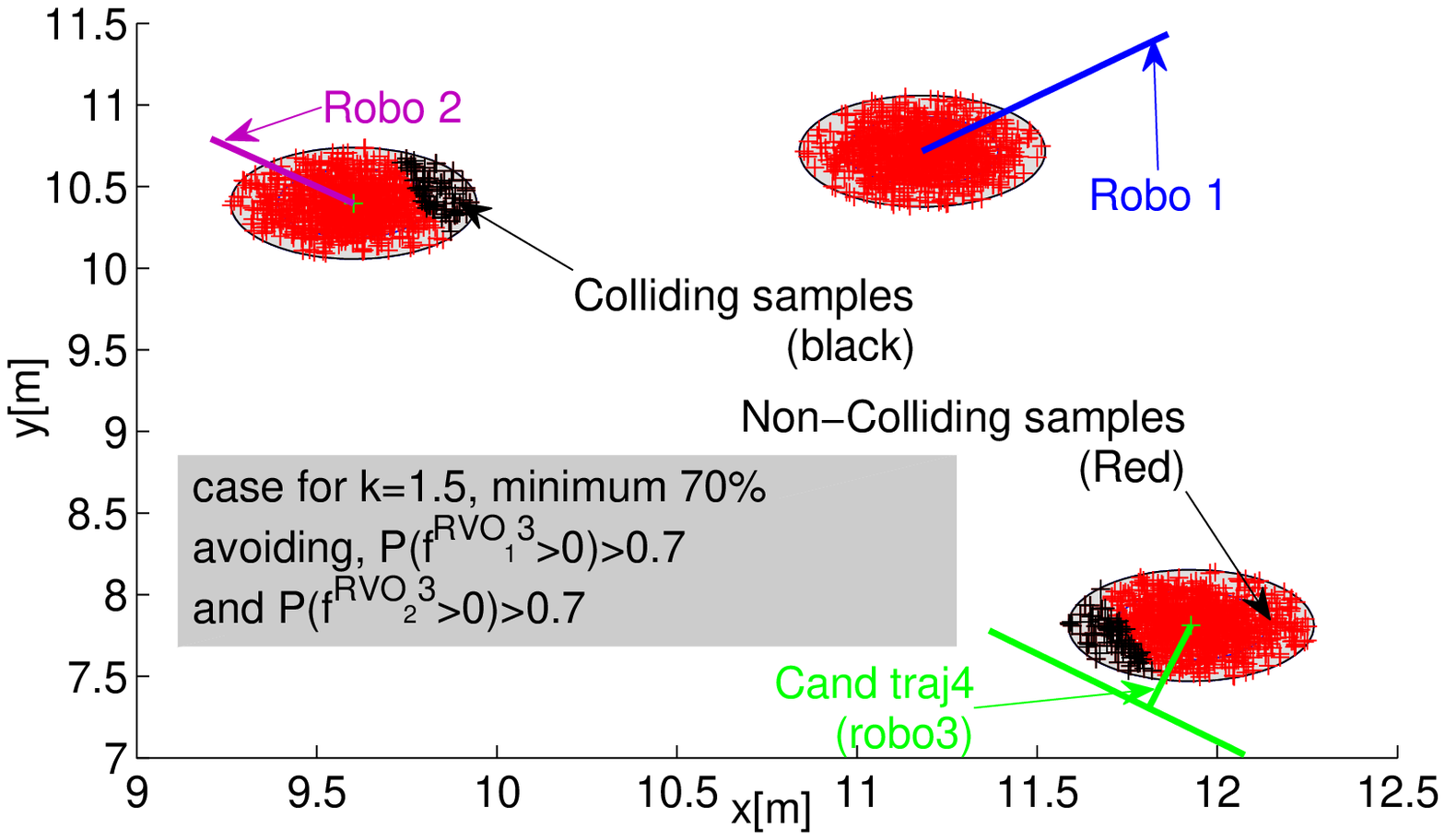}
        \label{k2rob3}
        }             
\caption{(a)-(c): Results are same as figure \ref{k1rob1}-\ref{k1rob3}, but are now obtained for $k=1.5$. With this value of $k$, the probability with which (\ref{PRVO}) should get satisfied for each robot should be 0.70, which as shown in figure is indeed the case. The validation was done through sampling procedure similar to figure 2.}
\end{figure*}

Figure \ref{3rob_coll} shows a collision scenario with three robots. In line with the time scaling based methodology described in section \ref{timescale}, each robot generates multiple candidate trajectories (figure \ref{candtraj1}) and then solves time scaled variant of the surrogate constraints (\ref{timescalevariant}) along them. Depending on the value of $k$, a particular candidate trajectory and its corresponding time scale is chosen. Figures, \ref{k1rob1}-\ref{k2rob3} summarizes these results. Consider, figure \ref{k1rob1}-\ref{k1rob3}, where each robot solves (\ref{expecvariance}) for $k=1$. According to Canetlli's based bounds, discussed in section 5.2, this would mean, that PRVO constraints (\ref{PRVO}) should be satisfied with atleast probability $0.5$. We validate this by sampling position and velocity samples from the uncertainty ellipses of the robots and evaluating what percentage of these samples satisfy the deterministic RVO constraints (\ref{RVO_comp}). As shown in the figures, the minimum probability observed agrees with Cantelli's bounds. Figures \ref{k2rob1}-\ref{k2rob3} solves (\ref{expecvariance}) for $k=1.5$ and as shown, the minimum probability with which PRVO is satisfied for each robot increases to 0.75, which is again in accordance with the Cantellis bounds.

\begin{figure*}[!h]
  \centering
    \subfigure[]{
\includegraphics[width= 4.0cm, height=4.3cm]{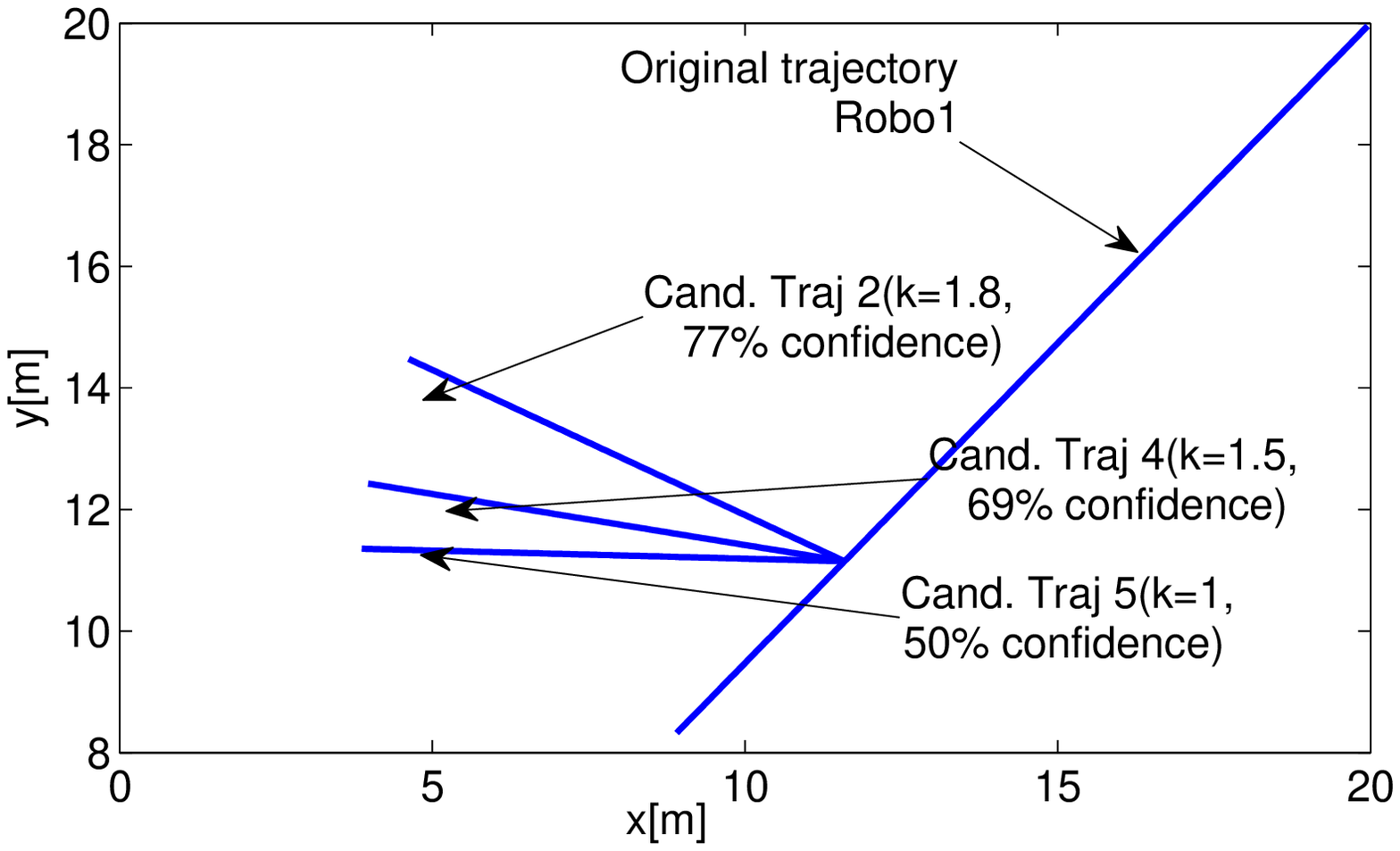}
        \label{candtrajrob1}
        }\hspace{-0.6cm}
\subfigure[]{
\includegraphics[width= 4.0cm, height=4.3cm]{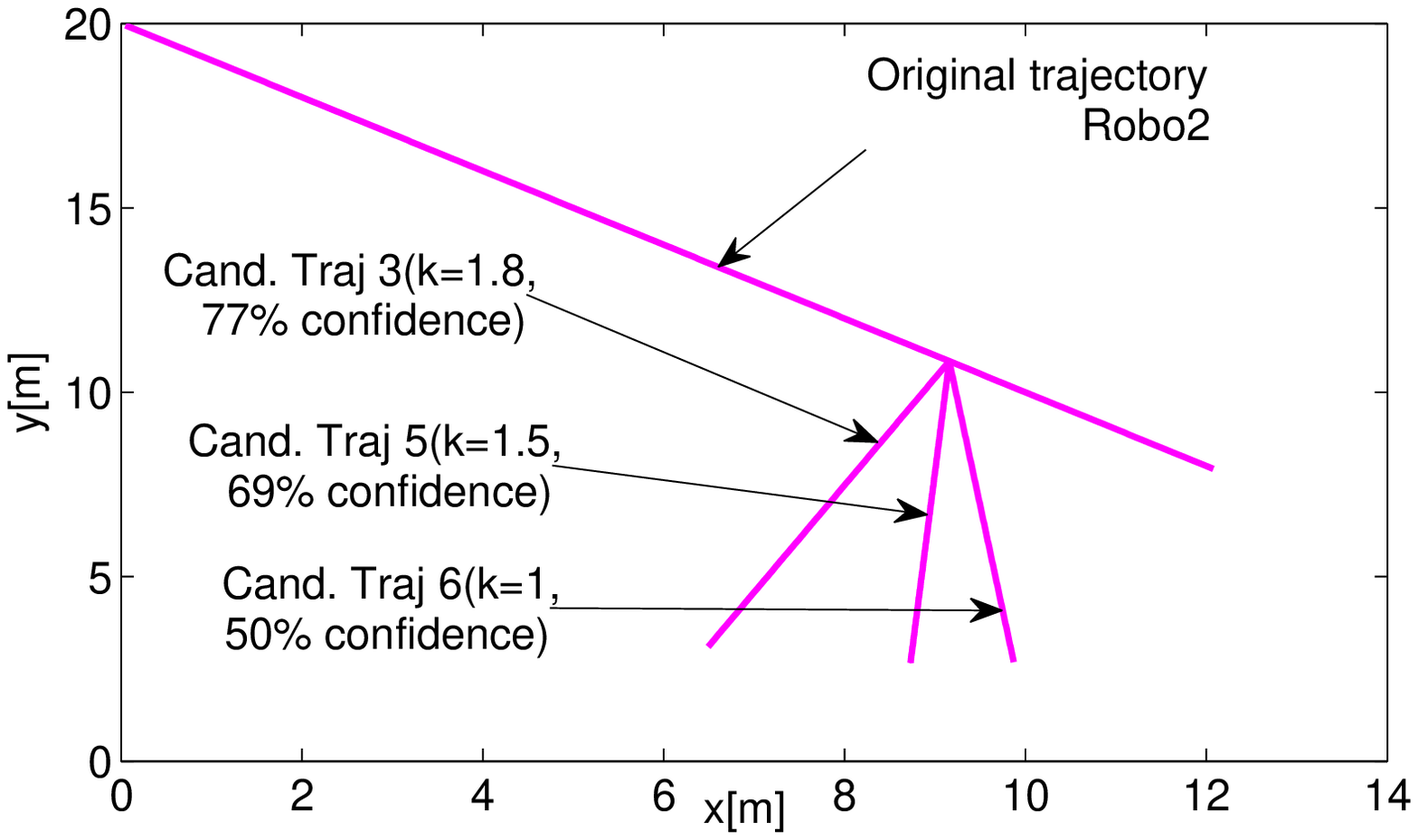}
        \label{candtrajrob2}
        }\hspace{-0.6cm}
        \subfigure[]{
\includegraphics[width= 4.0cm, height=4.3cm]{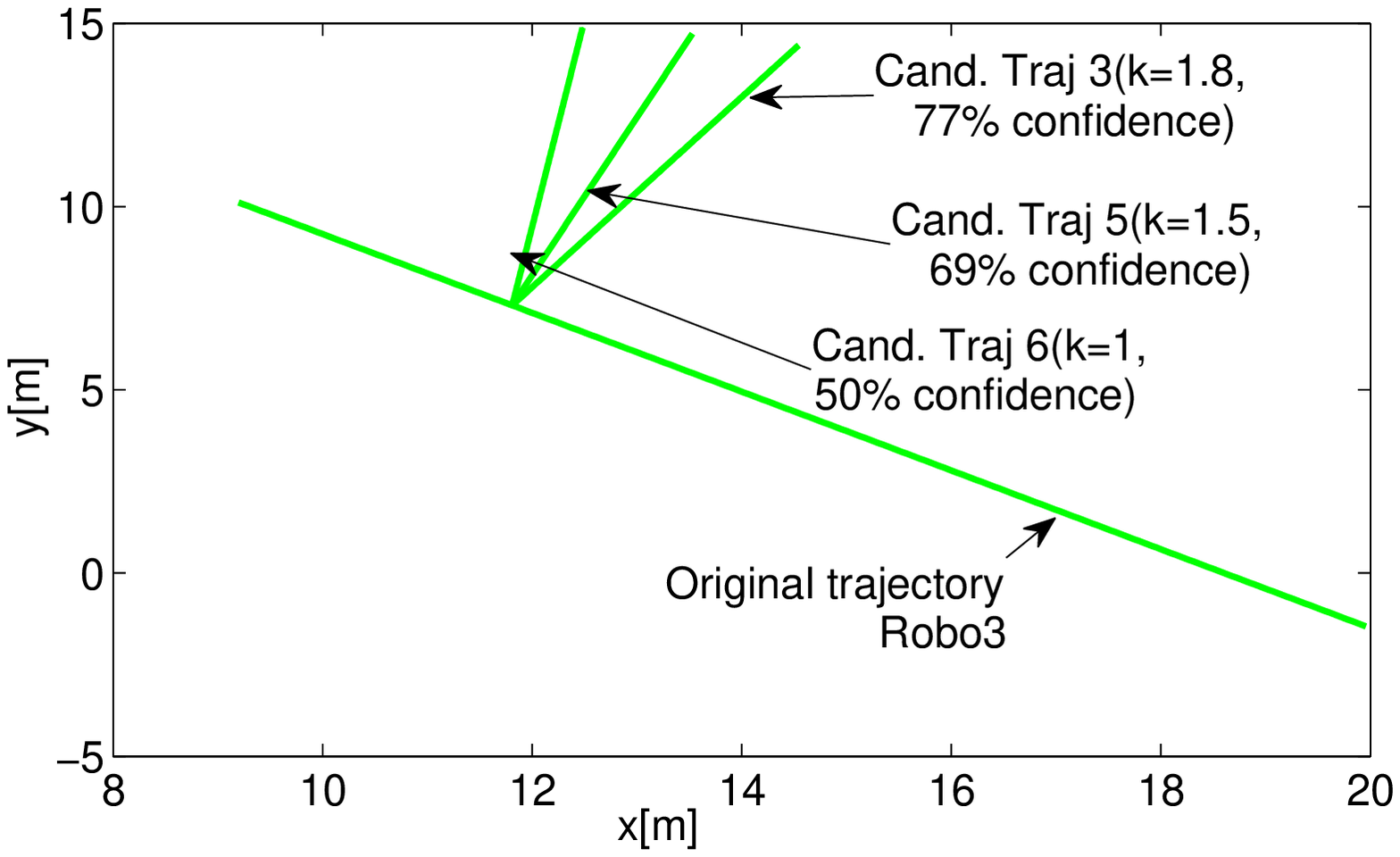}
        \label{candtrajrob3}
        }             
\caption{(a)-(c) Collision avoidance maneuvers for various values of $k$ or in other various probabilities $\eta$ of PRVO constraints (\ref{PRVO}). As can be seen, higher probabilities require each robot to take a larger deviation from the current trajectory.}
\end{figure*}

Figures \ref{candtrajrob1}-(\ref{candtrajrob3}) summarizes the avoidance maneuvers for various values of $k$ or in other words for various probabilities $\eta$. As shown, with increase in $k$, the deviation from the current trajectory increases for each robot.

 \subsection{Illustrating the need for incorporating motion(ego) uncertainty while computing collision free velocities }
In this section, we highlight that if self motion uncertainties of robots are ignored, then it is difficult to reliably infer, the probability with which the RVO constraints are actually satisfied with each robot. To put it alternatively, the Cantellis bounds based mapping between the solution space of the surrogate constraints (\ref{expecvariance}) and the PRVO (\ref{PRVO}) does not hold if self motion uncertainties of the robots are ignored. 

\begin{figure*}[!h]
  \centering
   \subfigure[]{
\includegraphics[width= 4.3cm, height=4.1cm]{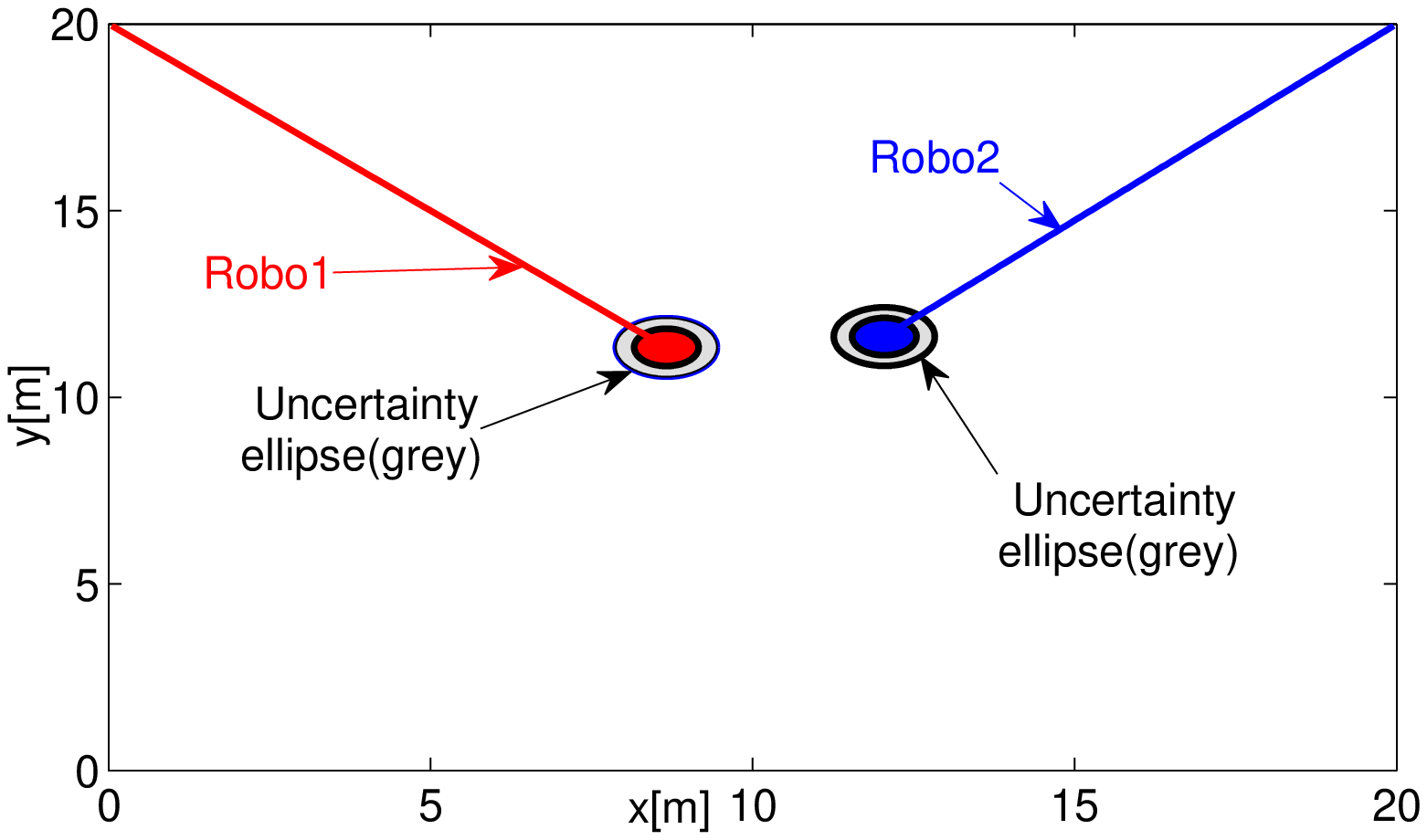}
        \label{2_robo_config_comparision}
        }\hspace{-0.6cm}
    \subfigure[]{
\includegraphics[width= 4.1cm, height=4.3cm]{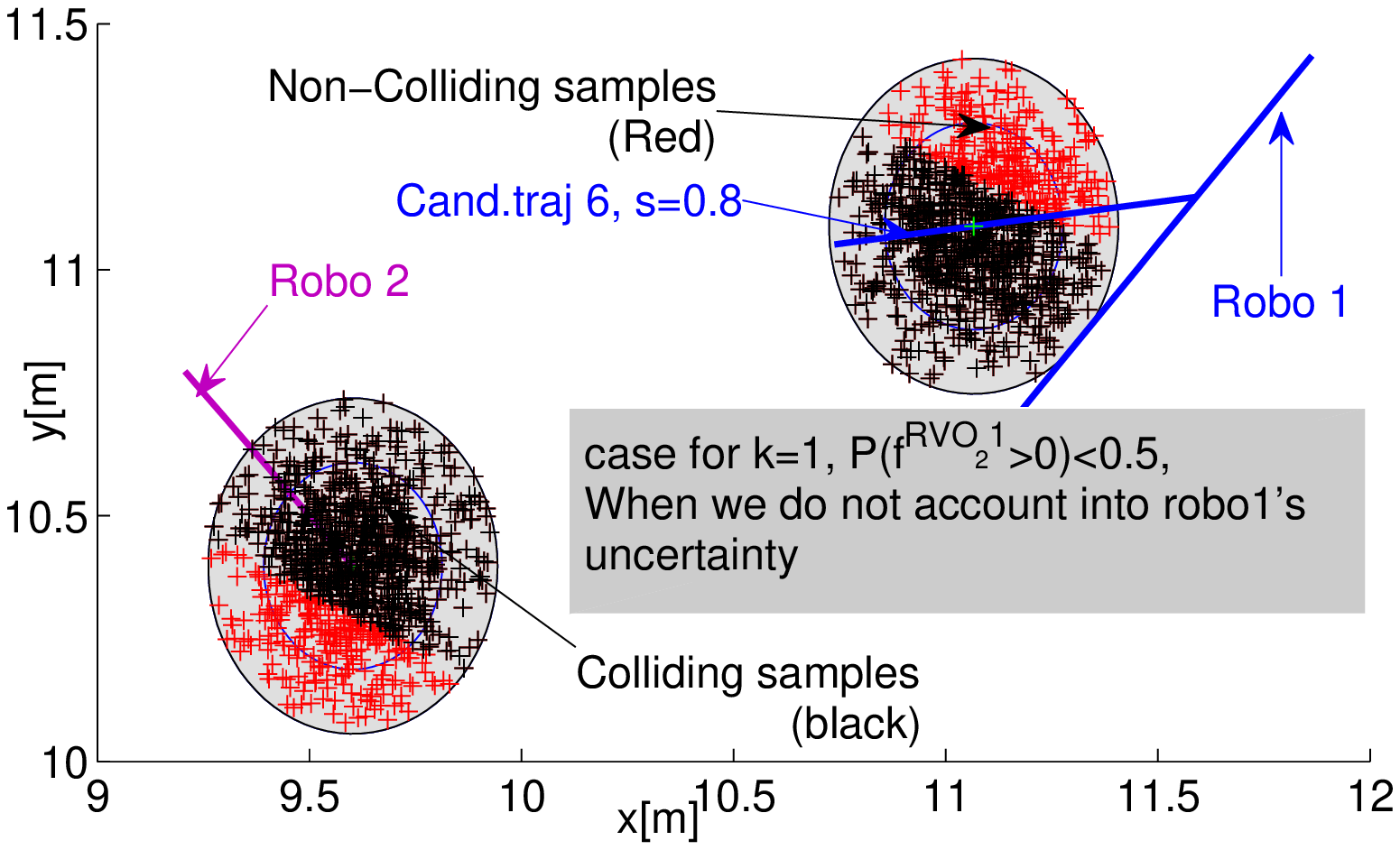}
        \label{robo1_without_robos_uncert}
        }\hspace{-0.6cm}
\subfigure[]{
\includegraphics[width= 4.1cm, height=4.3cm]{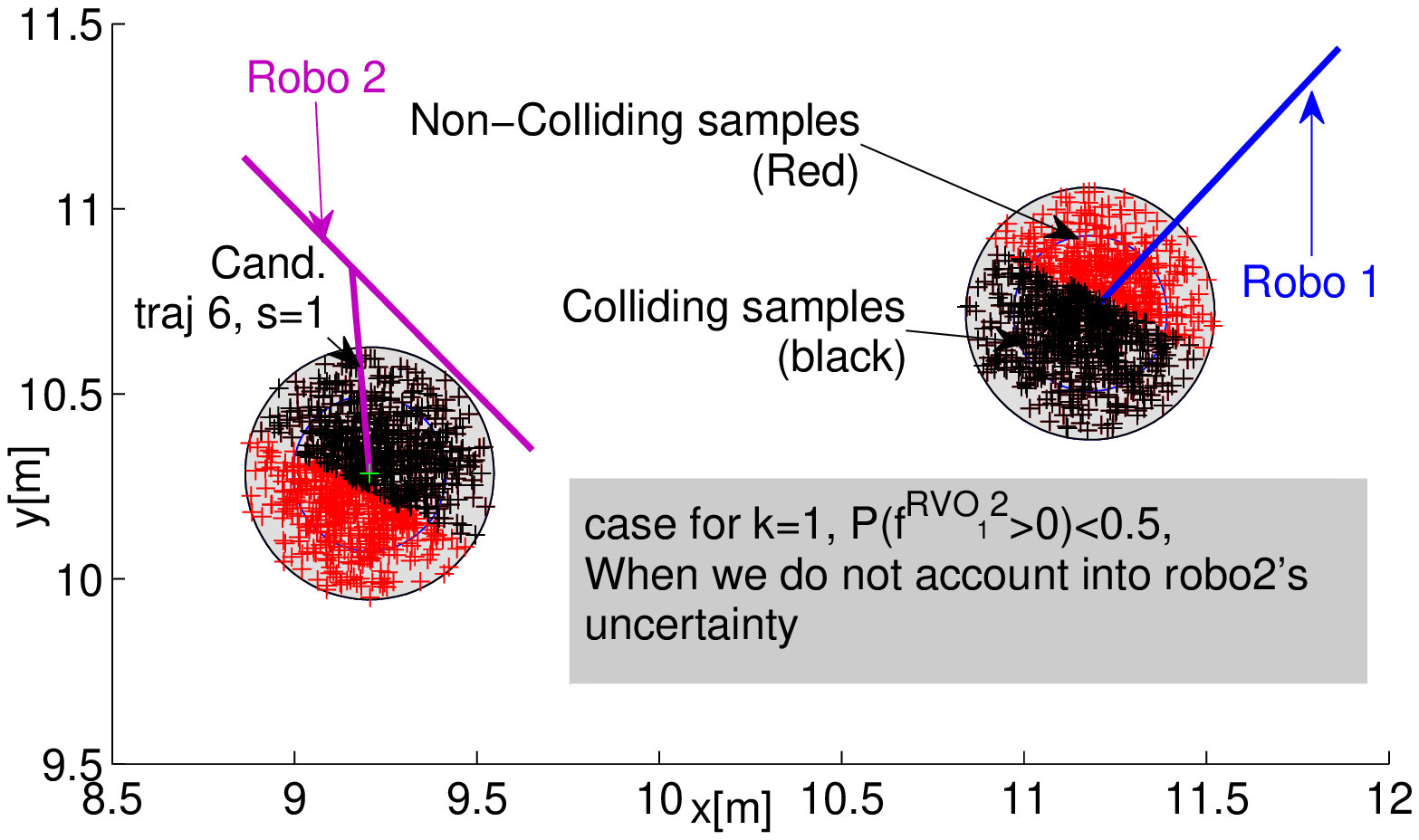}
        \label{robo2_without_robos_uncert}
        }\hspace{-0.6cm}
        \subfigure[]{
\includegraphics[width= 4.1cm, height=4.3cm]{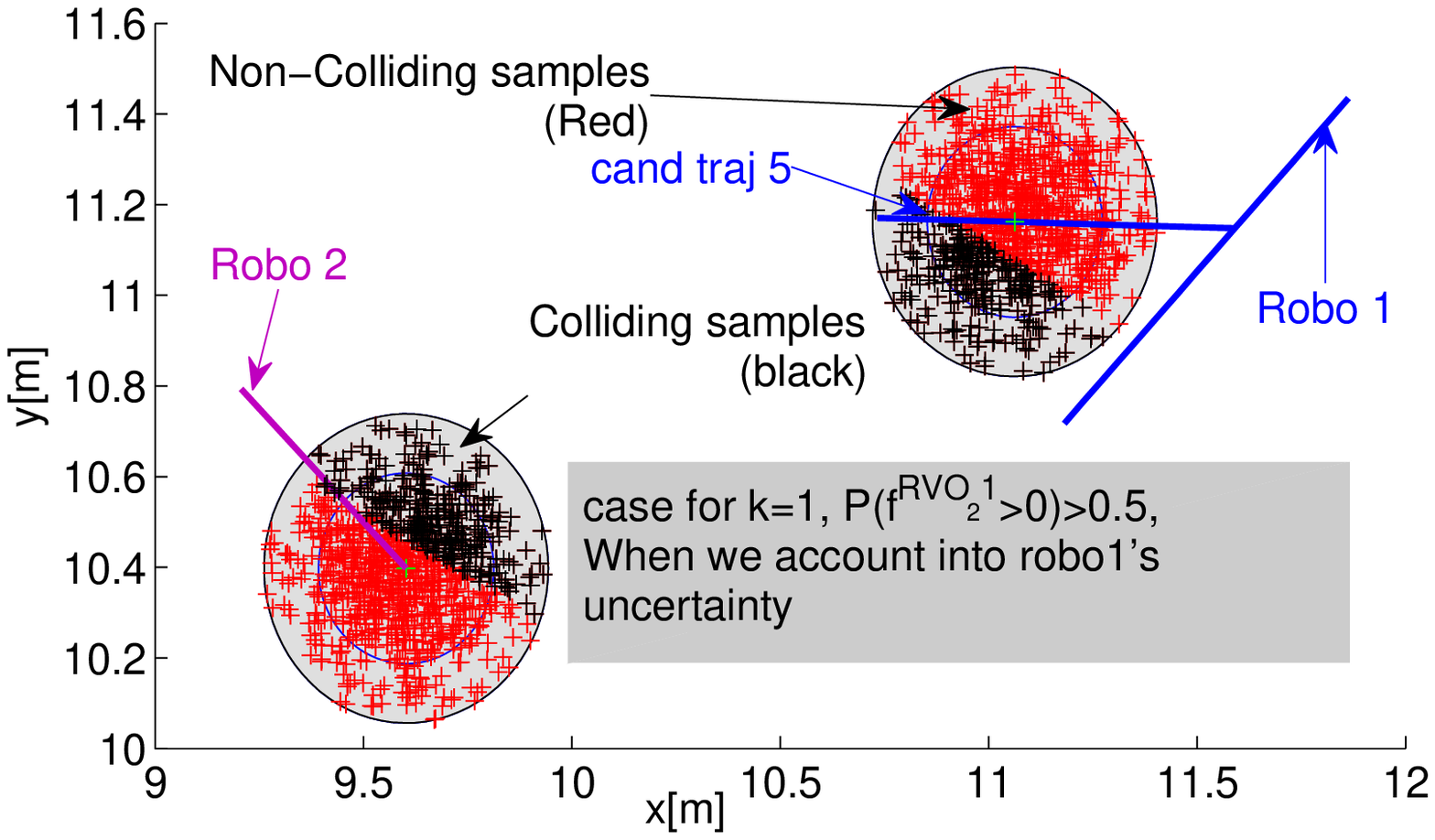}
        \label{robo1_with_robos_uncert}
        }\hspace{-0.6cm}
        \subfigure[]{
\includegraphics[width= 4.3cm, height= 4.3cm]{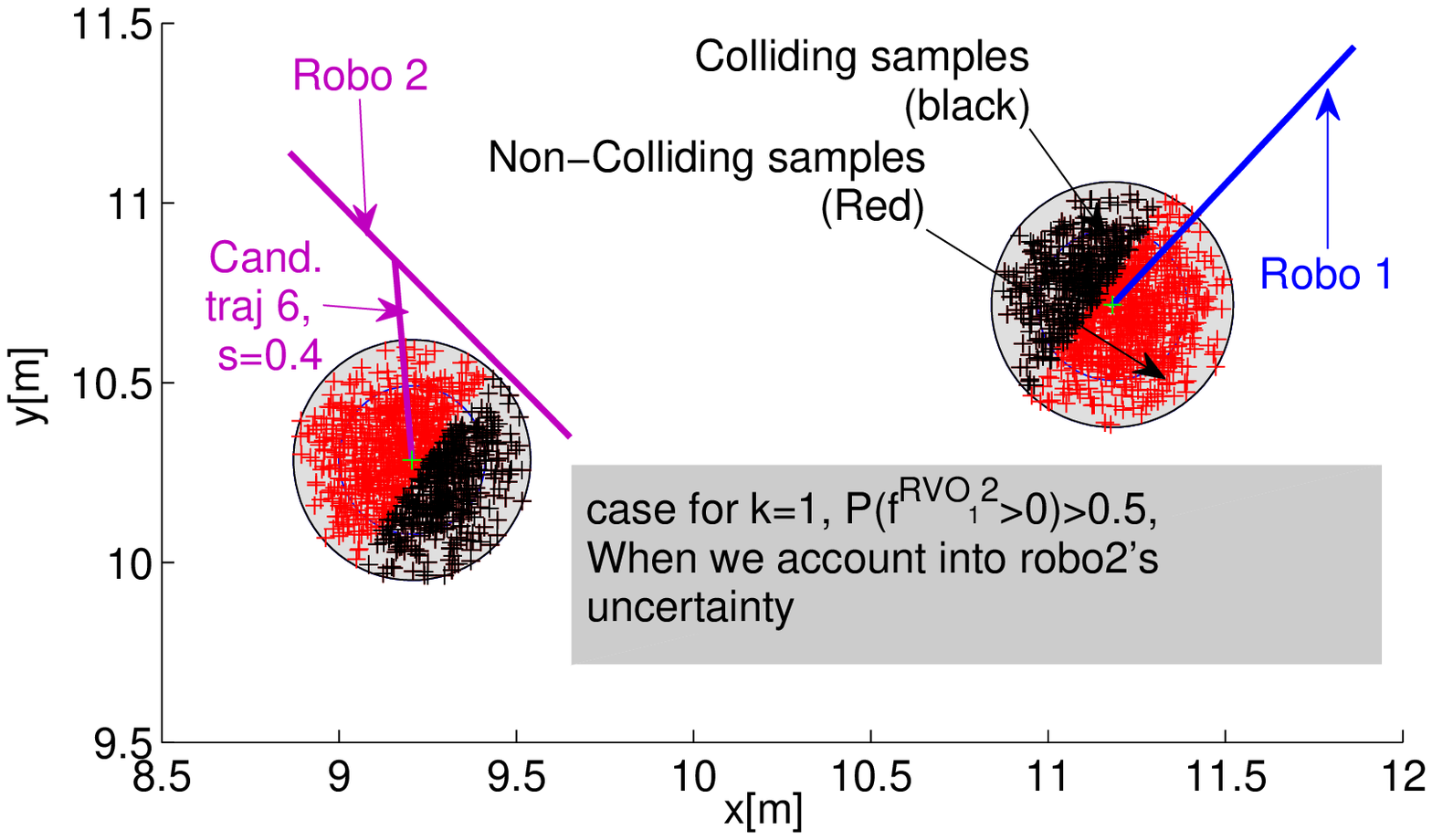}
\label{robo2_with_robos_uncert}
    }  
    \caption{(a)-(e),(b) and (c) illustrates the effect of not considering the effect of motion uncertainty while computing collision free velocities, it can be observed from the number of colliding samples(black) that for k=1,the PRVO constraints (\ref{PRVO}) are not satisfied with a lower bound as 0.5, while (d) and(e) illustrate the effect of considering motion uncertainty, it can be seen here that for k=1, the PRVO constraints (\ref{PRVO}) are satisfied with atleast probability 0.5    }
\end{figure*}

We start by deriving a variant of the surrogate constraints (\ref{expecvariance}) but without considering equation (\ref{motion_uncert}), i.e, $\textbf{v}^i_{rvo}$ is no longer treated as a random variable. Now for a configuration of robots shown in figure \ref{2_robo_config_comparision}, consider figures \ref{robo1_without_robos_uncert}-\ref{robo2_without_robos_uncert}, where this variant is solved by each robot for $k=1$. Based on Cantelli's bounds, this should translate to PRVO being satisfied with $\eta\geq 0.5$. However, as shown, when evaluated through sampling, $\eta$ turns out to be less than 0.5. In contrast, in figures \ref{robo1_with_robos_uncert}-(\ref{robo2_with_robos_uncert}), where self motion uncertainties are considered, Cantellis bounds hold perfectly. 

\subsection{Advantage of PRVO over existing works }

One of the simplest approach to model the probabilistic variant of RVO is to grow the radius of the robots by a value corresponding to a particular confidence region of the uncertainty ellipses . This procedure has been illustrated  in \cite{hrvo} and \cite{hrvo_prvo}.Table\ref{comparision_agent1} and Table\ref{comparision_agent2}  compares this method with our proposed formulation for two robots for a scenario similar to Figure \ref{2_robo_config_comparision}. The solution space of the timescaled variant \ref{timescalevariant} of the surrogate functions for a particular value of 'k' that satisfies a 68 percentage and 80 percentage confidence contour of the uncertainty ellipse in the position space was obtained.This is then compared with the solution space that is obtained by enlarging the robot's radius by 68 percentage and 80 percentage confidence contours.While we enlarge the radius by the desired confidence region, it is also important to take into consideration the probability of velocities that the robots can take. So the solution spaces presented in tables \ref{comparision_agent1} and \ref{comparision_agent2}  were evaluated for the most probable velocities (velocities that are very close to the mean velocities).  It is clearly seen that the robots , especially robot $1$ may tend to decelerate a bit higher if the solution space obtained by enlarging the radius is followed. 
\small
\begin{table*}[!h]
\caption{Comparison of the solution space obtained from the proposed formulation and that by the method of enlarging the robot's radius for agent 1}
\centering
\begin{tabular}{|c|c|c|}
\hline\hline
 Formulation & 68 $\%$ contour & 80 $\%$ contour\\
\hline
Expanding the radius by a desired confidence contour  & $[0\hspace{0.1cm}0.59]$ & $[0\hspace{0.1cm}0.4]$\\
\hline
Solution space of surrogate functions  & $[0\hspace{0.1cm}0.67]$ & $[0\hspace{0.1cm}0.5]$\\
\hline
\end{tabular}
\label{comparision_agent1}
\end{table*}
\normalsize
\vspace{-5mm}
\begin{table*}[!h]
\caption{Comparison of the solution space obtained from the proposed formulation and that by the method of enlarging the robot's radius for agent 2}
\centering
\begin{tabular}{|c|c|c| }
\hline\hline
 Formulation & 68 $\%$ contour & 80 $\%$ contour\\
\hline
Expanding the radius by a desired confidence contour  & $[0\hspace{0.1cm}0.35]\cup[1.1\hspace{0.1cm}\infty)$& $[0\hspace{0.1cm}0.23]\cup[1.1\hspace{0.1cm}\infty)$\\
\hline
Solution space of surrogate functions  & $[0\hspace{0.1cm}0.37]\cup[1.1\hspace{0.1cm}\infty)$& $[0\hspace{0.1cm}0.27]\cup[1.1\hspace{0.1cm}\infty)$\\
\hline
\end{tabular}
\label{comparision_agent2}

\end{table*}

\normalsize


%

\vspace{-5mm}
\section{Conclusions, Limitations and Future Work}            

\subsection{Conclusions}
In this paper, we have presented the probabilistic variant of RVO, defined as chance constraints \ref{PRVO} over the inequalities defined by the deterministic RVO. These chance constraints are generally computationally intractable,and the consideration of ego and estimation uncertainties further increases its complexity.This paper attempts to approximate such a chance constraint through a family of surrogate constraints \ref{expecvariance} that have a closed form characterization of their solution space. Further a closed form mapping based on cantelli's inequalities is provided that maps the solution space of these surrogates to the probability of the chance constraint being satisfied.

\subsection{Limitations}
The computed avoidance maneuvers are piece-wise straight line trajectories with no velocity continuity or acceleration bounds. For practical implementation, these needs to be incorporated. The cantelli's inequality \ref{Cantelli} represented here can act as a weak bound at times. Formulation of an efficient scheme for mapping the solution space of the surrogate functions to the probability of the chance constraint being satisfied is very important and would form the main crux of our future work.

%


\end{document}